# Estimating oil recovery factor using machine learning: Applications of XGBoost classification

Alireza Roustazadeh[1], Behzad Ghanbarian[1*], Frank Male[2], Mohammad B. Shadmand[3], Vahid Taslimitehrani[4], and Larry W. Lake[5]

[1] Porous Media Research Lab, Department of Geology, Kansas State University, Manhattan 66506 KS, United States

[2] Pennsylvania State University, University Park 16802 PA, United States

[3] Department of Electrical and Computer Engineering, College of Engineering, University of Illinois at Chicago, Chicago 60607 IL, United States

[4] Staff Machine Learning Scientist at Realtor.com, San Francisco CA, United States

[5] Hildebrand Department of Petroleum and Geosystems Engineering, University of Texas at Austin, Austin 78712 TX, United States

* Corresponding author's email address: ghanbarian@ksu.edu

**Abstract**

In petroleum engineering, it is essential to determine the ultimate recovery factor, RF, particularly before exploitation and exploration. However, accurately estimating requires data that is not necessarily available or measured at early stages of reservoir development. We, therefore, applied machine learning (ML), using readily available features, to estimate oil RF for ten classes defined in this study. To construct the ML models, we applied the XGBoost classification algorithm. Classification was chosen because recovery factor is bounded from 0 to 1, much like probability.

Three databases were merged, leaving us with four different combinations to first train and test the ML models and then further evaluate them using an independent database including unseen data. The cross-validation method with ten folds was applied on the training datasets to assess the effectiveness of the models. To evaluate the accuracy and reliability of the models, the accuracy, neighborhood accuracy, and macro averaged f1 score were determined. Overall, results showed that the XGBoost classification algorithm could estimate the RF class with reasonable accuracies as high as 0.49 in the training datasets, 0.34 in the testing datasets and 0.2 in the independent databases used. We found that the reliability of the XGBoost model depended on the data in the training dataset meaning that the ML models were database dependent. The feature importance analysis and the SHAP approach showed that the most important features were reserves and reservoir area and thickness.

## 1. Introduction

Accurate estimation of ultimate recovery factor (RF) has broad applications to oil and gas exploration. The RF is an indicator of how feasible a reservoir would economically be produced during its lifetime (Beaumont et al., 2019). Accordingly, various methods, such as dynamic reservoir simulations, production decline curve analysis, material balance, and field analogues (Aliyuda and Howell, 2019; Maselugbo et al., 2017; Omoniyi and Adeolu, 2014) were presented in the past several decades to determine the RF from available measured data. Such methods, however, either are computationally demanding, associated with significant uncertainties and errors, or require numerous input data that are not necessarily available before field development (Lin et al., 2012; Ling et al., 2013; Muskat and Woods, 1944; Parish et al., 1993). With recent advances in artificial intelligence, machine learning (ML) algorithms have been proactively used, particularly in the industry in the past several years (Tahmasebi et al., 2020). The general goal is training ML models to estimate reservoir properties e.g., RF from readily available or easily measurable data, which leads to have a robust and cost-efficient platform, especially with practical applications to industry (Holdaway, 2014; Mohaghegh and U, 2000; Srivastava et al., 2016).

In the literature, most ML models developed to estimate RF are based on regression analysis (Ahmadi and Chen, 2019; Aliyuda et al., 2020; Aliyuda and Howell, 2019; Alpak et al., 2019; Lee and Lake, 2015; Sharma et al., 2010; Tewari et al., 2019). For example, Srivastava et al. (2016) used machine learning on the 2013 version of Atlas of Gulf of Mexico database to classify different reservoirs. They conducted a principal component analysis on original features that they had in their database to reduce the dimensionality of features. Then, they used k-means clustering to group similar reservoirs together and, finally correlated input features to RF through partial least squares regression. Those authors found that k-means clustering in combination with the principal component analysis and partial least square regression led to unsatisfactory results, achieving a maximum Pearson's correlation coefficient of 0.2 between predictions and realized RF. A different approach, grouping reservoirs based on dimensionless numbers, such as gravity number, aspect ratio, and density number and using such numbers as input features to their partial least squares regression model resulted in better performance. The partial least squares regression and k-means clustering based on dimensionless numbers method yielded $R^2$ values as high as 0.92 and as low as 0.1 in 4 different clusters.

Using data from three different databases (TORIS, Digital Knowledge System, and the Oil and Gas Journal), Kaczmarczyk et al. (2014) applied clustering analysis in combination with classification tree approach to estimate primary, secondary, and tertiary RFs. Their decision tree model was constructed using six different reservoir and fluid properties (i.e., pressure, permeability, viscosity, porosity, API gravity, and depth).

Gupta et al. (2016) used the partial least squares regression method and data from Gulf of Mexico deep offshore assets to estimate RF variance between early appraisal and post sanction phase. Their work does not directly estimate RF. Instead, it forecasts the difference in predicted RF from two different stages in the development process.

In another study, Karacan (2020) used reservoir data collected from only 24 reservoirs in the United States and RFs from literature survey. He applied the fuzzy logic approach to estimate miscible $CO_2$-EOR RF from nine features, such as API, HCPV, porosity, permeability, depth, and initial pressure. Karacan (2020) reported the correlation coefficient of 0.94 for his fuzzy logic model with mean square error of 4%. However, it is difficult to draw statistical conclusions from such a small dataset.

Recently, Roustazadeh (2022) constructed regression-based models using three ML algorithms including XGBoost, support vector machines, and forward stepwise multiple regression. He found that the XGBoost regression model was slightly more accurate than the other two models. Roustazadeh (2022) also found database dependence in regression-based ML models, meaning that ML models are more accurate if testing and training data are statistically similar.

Classification-based ML models, on the other hand, are rarely used for estimating RF at the reservoir scale. As shown above, regression algorithms have been previously applied in the literature (Chen et al., 2022; Han and Bian, 2018; Kalam et al., 2022; Makhotin et al., 2021). Furthermore, most ML models are not further evaluated with new and unseen data not used in their training.

The main objectives of this study are to: (1) apply the extreme gradient boost (XGBoost) classification algorithm to construct ML models using large databases, (2) estimate the ultimate oil RF at the early stages, and (3) investigate the database dependence of classification-based ML models by evaluating them using unseen data.

## 2. Materials and Methods

In this section, we first explain data used in this study. Next, we describe how the data were prepared to construct the machine learning-based models using the XGBoost classification algorithm. After that, we describe how we use SHAP to the determine feature importance.

### 2.1. Databases

The data used in this study are from three different databases described briefly in the following.

**- Commercial database**

The private, commercial database includes more than 1200 conventional reservoirs (observations) with different rock and fluid properties from around the world. In this database, there are more than 200 features, such as pressure, water saturation, and porosity, reported for each reservoir. The commercial database has data for both oil and gas reservoirs. After filtering, we used data from 600 oil fields in this study. For further detail about the database, we refer the interested reader to Lee and Lake (2015).

**- Total Oil Recovery Information System (TORIS) database**

Gathered by National Petroleum Council and used by US Department of Energy, TORIS is a venerable database with information about the oil reservoirs of the United States of America. It has more than 1300 observations and 60 features but a relatively high number of missing values. For further information about this database, see Long (2016).

**- Bureau of Ocean Energy Management Atlas of Gulf of Mexico**

Bureau of Ocean Energy Management (BOEM) annually collects oil and gas data from conventional fields in the outer shelf of the Gulf of Mexico (Burgess, Grant L.; Cross, Kellie K.; Kazanis, 2019). In this study, we used the 2018 version of this database (simply called Atlas hereafter). The Atlas database consists of more than 13,000 observations and 80 features for both oil and gas reservoirs. Only the data from oil fields were used to construct ML models. It is the most complete database (with fewer missing values) among the three databases analyzed in this study.

Duplicate observations were detected and removed to assure no reservoir was present in more than one database. The observation with fewer input features was removed. Figure 1 shows the distributions of the RF, porosity ($ft^3/ft^3$), and natural logarithm of permeability in the Commercial, Atlas and TORIS databases. As can be seen, the distributions of RF, permeability, and porosity in the TORIS and Commercial databases are similar, while the Atlas database is different. The distribution of natural logarithm of permeability is left skewed and heavy tailed for the TORIS and Commercial databases, while slightly left tailed for the Atlas database. In contrast, the RF distributions for all the three databases are right skewed with heavier tails in the TORIS and Commercial databases.

Figure 2 shows the box and whisker plots for oil RF, porosity ($ft^3/ft^3$), and natural logarithm of permeability in each database used in this study. As can be seen, the databases have a few outliers when it comes to the important reservoir quality attributes e.g., porosity, permeability, and RF values. It can also be seen that the TORIS and Commercial databases have similar interquartile ranges, median, and mean values. The Atlas database, however, consists of samples with smaller ranges in porosity and permeability compared to those in the TORIS and Commercial databases. It also has considerably different mean and median values than TORIS and Commercial shown in all three plots. TORIS had the greatest number of RF outlier values, while Atlas had the fewest among the three databases. Note that most of the oil RF values in all three databases range between 0.2 and 0.45.

The Commercial, TORIS, and Atlas described above were used to construct the classification models to estimate the oil RF from other properties. These databases were merged using four different combinations to create larger databases. More specifically, the combined databases TORIS and Commercial was named TC, TORIS and Atlas named TA, and Commercial and Atlas called CA. The last combination was created by merging all the three databases i.e., TORIS, Commercial, and Atlas named TCA. In the first three combinations (i.e., TC, TA, and CA), the third database was used for testing.

## 2.2. Data preparation

Data preparation started with us removing any reservoir in any database that did not have an RF value. Although many different geological, petrophysical, and production features were available, we selected only those features that were in all three databases. Moreover, any feature associated with post exploration phase data (e.g., production time, final abandonment pressure, and cumulative production) was removed. To conserve the data similarity across all databases, the selected features were constrained to commonly expected values. For example, the oil formation volume factor was restricted between 1 and 3 ($1 \leq B_O \leq 3$), gas oil ratio from 0 to 60 ($0 \leq GOR \leq 60$), and reserves between 0 and $5 \times 10^{11}$ stock tank barrels. Any feature with more than 70% missing values and any reservoir with more than 55% missing values were removed. Table 1 lists the final input features and target variable (RF) as well as their ranges in each database. It should be mentioned that the Atlas database defines reserves as the hydrocarbon remaining that could be economically recovered.

For the purpose of constructing ML classification models, ten RF classes with intervals of 0.1 were defined. For example, samples with the RF values ranged between 0 and 0.1 were grouped in class 0. Similarly, samples with $0.9 \leq RF1$ in class 9. We randomly selected 90% of the data for training and cross-validation and 10% for testing, similar to the study by Dias et al. (2020). The purpose of training and testing splits is to provide the ML models with enough training data to be constructed upon and avoid over-fitting (Agbadze et al., 2022). We used 10-fold cross-validation to tune the model hyperparameters on the training dataset (Carpenter, 2021). We split the data into the training and testing datasets before any data preprocessing and imputation. Then, we preproccessed the testing data with the identical parameters used on the training dataset (Kapoor and Narayanan, 2021). We did not perform any imputation on the independent databases. Instead, we removed samples with missing data either input features or target variable. This is because the independent databases were used to further evaluate the ML models using real-time data, not imputed ones. As stated earlier, observations in the training and testing datasets were de-duplicated. To ensure the fraction of samples in each class for the training and testing datasets was approximately equal, both the training and testing datasets were stratified by recovery factor class (Anifowose et al., 2011; Helmy and Fatai, 2010).

To estimate missing values, one may use an imputation approach (Lin and Tsai, 2020). In this study, we developed an in-house imputation method. First, the data were sorted ascendingly based on their numerical RF values. Next, an array of 10 entries of the feature with missing values was selected. If the array had no missing value, the next array including 10 samples was selected.

If the total number of missing values to the total number of samples in the array was less than or equal to 10%, the modal value of the feature within that array was used to replace the missing values. If the number of missing values in the array was more than 10%, then more entries from the same feature was added to the array until the ratio of the missing values to the total number of samples equaled or dropped below 10%.

To minimize bias and avoid prioritizing features with larger magnitudes, data should be standardized and normalized (Tunkiel et al., 2022). We used the Gaussian rank transformation for standardizing the data. Using the Gaussian rank transformation, the distribution of the data was transformed to approximately Gaussian by ranking the data from the smallest to the largest value and calculating its inverse error function. For normalizing, the data were transformed to unit magnitude (between 0 and 1) using the formula $X_{norm} = (x-x_{min})/(x_{max}-x_{min})$ (Pedregosa et al., 2011).

## 2.3. Machine learning and XGBoost classification model

The aim of this study was applying a ML classification model to estimate an expected RF interval for a reservoir at the exploration stage. We used the XGBoost model, an open-access source algorithm compatible with different programming languages e.g., Python. XGBoost may be used for either classification (He et al., 2022; Pirizadeh et al., 2021) or regression (Dong et al., 2022; Pan et al., 2022) problems. It is an ensemble ML algorithm with roots in decision tree based on gradient boosting (Chen and Guestrin, 2016). In the XGBoost framework, weak learners are stacked to create strong learners in sequential steps (Tang et al., 2021).
XGBoost can provide more accurate estimations than other ML models, such as support vector machine (Liu et al., 2019; Pirizadeh et al., 2021), random forest (He et al., 2022; Zhao et al., 2022), and k-nearest neighbor (Chen et al., 2021; He et al., 2022). In addition to that, gradient boosting algorithms (e.g., XGBoost) are unaffected by collinearity among input features as well as between input features and target variables (Kotsiantis, 2013).

During hyperparameter tuning, we optimized sets of two hyperparameters simultaneously, while keeping the rest of them constant. This process iterated until all hyperparameters were tuned was done on the training dataset with 10-fold cross-validation and using the multiclass logistic (logarithmic) loss function (mlogloss). The values of hyperparameters yielded the lowest value of mlogloss were selected as the final values for constructing the ML models.

## 2.4. Model evaluation

The performance and accuracy of the models were evaluated using three parameters including accuracy, neighborhood accuracy, and macro averaged f1 score given by

$$Accuracy = \frac{number\ of\ correct\ estimations}{total\ number\ of\ estimations} \tag{1}$$

$$Neighborhood\ accuracy = \frac{number\ of\ estimations \in neighbor\ classes}{total\ number\ of\ estimations} \tag{2}$$

$$f1\ score = \frac{\sum_{i=1}^{n} class_i's\ f1\ score}{total\ number\ of\ classes} \tag{3}$$

Where *n* is the number of classes. All these parameters range from 0 and 1. Neighborhood accuracy indicates the number of estimations in one class above and/or below the desired class.

## 2.5. Feature importance: Shapley Additive exPlanations

Shapley Additive exPlanations, commonly known as SHAP, can estimate the importance of each feature and its impact on ML model outputs. It was originally proposed by Shapley (1953) who applied concepts from game theory to determine the outcome of a game based on every individual player's input. It explains how each feature and its data aid an algorithm to construct patterns between input features and target variables and make estimations (Lundberg and Lee, 2017). SHAP has been widely applied to determine feature importance (Kong et al., 2021; Li et al., 2022; Male et al., 2020).

## 2.6. Workflow of the constructed models

Figure 3 presents the workflow to construct the ML models on the databases used in this study including TC (TORIS merges with Commercial), TA (TORIS merged with Atlas), CA (Commercial merged with Atlas), and TCA (the combination of all three databases). Every model was constructed using XGBoost to estimate the oil RF classes in the training and testing datasets as well as independent databases. In total, four ML models were constructed and evaluated using accuracy, neighborhood accuracy, and the macro average f1 score.

## 3. Results

### 3.1. Hyperparameter tuning

Table 2 lists the optimized hyperparameters used to construct the classification-based ML models using the XGBoost algorithm. As the number of samples increased in the training datasets (from 1502 samples in the TC database to 5,461 samples in the TCA database), the hyperparameters were more easily optimized. The optimum learning rate decreased as the training dataset became larger. This means that the best performing models took more steps to reach the optimum value that had the minimum overfitting or underfitting.

### 3.2. Oil RF estimation

Table 3 presents the calculated accuracy, neighborhood accuracy, and macro averaged f1 score for the training and testing datasets as well as the independent databases used to further evaluate the constructed ML models. The number of samples in the training datasets were 1502 for TC, 4779 for TA, 4654 for CA, and 5461 for TCA. The lowest accuracy and f1 score values (i.e., 0.32 and 0.11) belong to the database TC, which had the least number of observations. An accuracy of 0.32 means that the XGBoost model predicted the correct oil RF class for 32% of the samples. The neighborhood accuracy of 0.37 indicates that for 37% of the samples, the XGBoost model estimated the RF either in the class below or above (i.e., within the two classes around) the desired class. The f1 score of 0.11 reported in Table 3 represents that the model had an unsatisfactory performance on the database TC (Gu et al., 2021; Li et al., 2022). Worse performance of the XGBoost algorithm for the database TC could be due to the limited number of samples in this database (1502 samples in the training dataset).

As the number of samples in the training dataset increases from the database TC to TA, the accuracy and f1 score values also increase (accuracy = 0.32 vs. 0.52 and f1 score = 0.11 vs. 0.35). Similar results were observed for the databases CA and TCA, which indicates the ML models are database dependent. Although the database TCA has nearly 700 samples more than the database TA, its accuracy and f1 score values (0.49 and 0.33) are not greatly different from those reported for TA (0.52 and 0.35). However, the RF estimations for the databases TCA and TA are more accurate than those for the database CA in the training. We found similar results for the testing datasets. The accuracy and f1 score values are very similar for the databases TA, CA, and TCA, while greater than those obtained for the TC database. We found the testing f1 score values slightly less than the training values for the TC and CA databases, while the difference in f1 scores is more significant for the TA and TCA databases.

The accuracy of estimating the RF within the correct class or one class below or above the correct class can be determined by adding the accuracy and neighborhood accuracy values reported in Table 3. For the training dataset, we found the total accuracy of 0.69, 0.78, 0.75, and 0.77 for the databases TC, TA, CA, and TCA, respectively. For the testing dataset, similar values were found (0.71, 0.71, 0.73, and 0.71 for the TC, TA, CA, and TCA). These values show that the constructed models estimate the oil RF with nearly 70% accuracy within the correct or a neighboring class. Given that the interval in the RF classes is 0.1 (10%), the obtained results indicate reasonable estimations by the XGBoost model.

As Table 3 shows, the classification-based ML models could not estimate the RF for the independent databases as accurate as that for the testing datasets. For the database TC, the accuracy for the test dataset is 0.26, which dropped down to 0.20 for the independent database. Similarly, the accuracy decreased from 0.34 to 0.18 for the TA database and from 0.33 to 0.18 for the CA database. In the TCA combination all the three databases merged and split to training and testing datasets, so there is no independent database to further evaluate the constructed model. The low f1 scores listed for the independent databases (i.e., 0.06, 0.09, and 0.08) demonstrate that the ML models are database-dependent. We also calculated the total accuracy of oil RF estimations by adding the accuracy and neighbor accuracy and found 0.59 for the database TC, 0.52 for the database TA, and 0.51 for the database CA. The models estimate the oil RF with almost 55% accuracy within the correct class or one class below or above it.

Figure 4 presents bubble plots showing the estimated versus actual oil RF in each class for the four database combinations. The size of bubbles denotes the number of samples indicated on each bubble, and the solid line represents the 1:1 line. Large bubbles around the 1:1 line show accurate estimations for more observations. For the training plots, the large bubbles are mainly around the 1:1 line. However, for the testing plots, the constructed models tend to overestimate the oil RF for lower classes and to underestimate it for higher classes, in accord with the results of Lee and Lake (2015), Makhotin et al.(2021), and Roustazadeh (2022). Figure 4 shows that most samples have an RF value between 0.2 and 0.5 corresponding to intermediate classes (*i.e.*, 4, 5, and 6).

Figure 4 also shows that most training and testing observations are either on the 1:1 line or right below or above it. This is consistent with high total accuracies ($> 0.70$). Particularly for $0.2 < RF < 0.5$, the performance of the constructed ML models is satisfactory. This range corresponds to most conventional oil reservoirs (Fetkovich et al., 1996; Lake et al., 2014), indicating the potential practical applications to such types of reservoirs. Conventional reservoirs have an average RF of about 0.35 (Sheng, 2013).

### 3.3. Feature importance

Feature importance analysis was performed for the four combinations of databases studied here. Figure 5 shows the results of SHAP feature importance analysis. As can be observed, reserves and area are among the top features in the ML models constructed based on the databases TA, CA, and TCA. Their feature importance analyses in general are similar. The database TC, however, has a different feature importance selection. We found permeability, water saturation, and API gravity were the top three features for predictions made on the TC database. This may be due to the TC database being more sparsely populated.

Figure 5 also shows how much each of the features have contributed to prediction of each class. All models used most or all of the features to make predictions for the mid-range RF classes e.g., class 2, class 3, class 4, and class 5. As we move to rarer RF classes, the models used fewer features to make predictions. This could show how a larger number of samples within the mid-range RF classes affected the predictions of the models.

**4. Discussion**

The classification-based ML models constructed in this study showed satisfactory performances for the training and testing datasets. However, the performance of the models for the independent databases were not as satisfactory as for the testing datasets. Schaap and Leij (1998) are first who demonstrated that regression-based ML models are database-dependent meaning that they may not provide satisfactory estimations, if evaluated using new and unseen data not used to train them. The database dependence of ML models was recently highlighted by Ghanbarian and Pachepsky (2022).

As we stated earlier, ML models are not typically assessed using unseen data and independent databases. Schaap and Leij (1998) are among the first who demonstrated the database dependent accuracy and reliability of ML models. The database dependence of regression-based ML models in the estimation of oil and gas RFs was recently addressed by Roustazadeh (2022) who showed that the accuracy and reliability of ML models depended on data and size of databases used to train them.

It is not straightforward to compare results of classification-based models with those of regression-based ones because evaluation parameters used in classification problems (*e.g.*, accuracy and f1 score) are different from those used in regression cases (*e.g.*, root mean square error and correlation coefficient). In the study by Roustazadeh (2022), the accuracy was quantified by root mean square error, correlation coefficient, and coefficient of determination for independent databases was substantially less than that for training and testing datasets. In this work, although the performance of the classification-based ML models for the independent databases is not as satisfactory that for the training and testing datasets, the difference is not substantial, and the ML models still provide reasonable oil RF estimations.

Our feature importance results for the TC database are consistent with those reported by Makhotin et al. (2022) who applied the regression-based gradient boosting approach to estimate the oil RF. Those authors constructed two models based on pre- and post-production data using two databases i.e., TORIS, composed on 1381 oil reservoirs from the United States, and Proprietary, containing 1119 oil reservoirs from around the world collected by Russian oil companies. Result of their pre-production model showed that the top four features detected using the F-score were permeability, water saturation, viscosity, and API gravity (see their Fig. 7), and similar to our results obtained for the TC database (Fig. 5). Although Makhotin et al. (2022) found the reservoir thickness among the top five features, reserves was among the least impactful features in their pre-production analysis, again consistent with our results presented in Fig. 5 and the database TC.

## 5. Conclusion

In this study, we constructed classification-based ML models to estimate the oil RF from readily available data at early stages, such as exploration and exploitation. We collected thousands of reservoir observations by combining three databases. To address whether the classification-based ML models are database-dependent, we used four combinations of the three databases. The oil RF data were grouped into ten classes with 10% intervals.

Using the XGBoost classification algorithm, we trained the ML models for each combination. Results showed that the constructed models were accurately trained for all the combinations except the one that had the lowest number of samples. The oil RF estimations for the testing datasets were within either the correct or a neighboring class 70% of the time. There was degradation between training and testing datasets. Models were tested on independent datasets to assess the database dependence of the ML models. Results showed that the accuracy of the models for the independent databases was less than that for the testing datasets. Further investigations are required to improve the accuracy and reliability of machine learning models.

**Acknowledgement**

The authors are grateful to Amirhossein Yazdavar, United Healthcare INC, for fruitful discussions and comments. Alireza Roustazadeh acknowledges Department of Geology, Kansas State University, for financial supports through William J. Barret Fund for Excellence in Geology and Gary & Kathie Sandlin Geology Scholarship. Behzad Ghanbarian acknowledges Kansas State University for support through faculty start-up fund. The Python codes as well as means, standard deviations, and distributions of input features used in this study are available at github.com/alirezaro93/Estimating-oil-recovery-factor-at-exploration-stage-using-XGBoost-classification-.

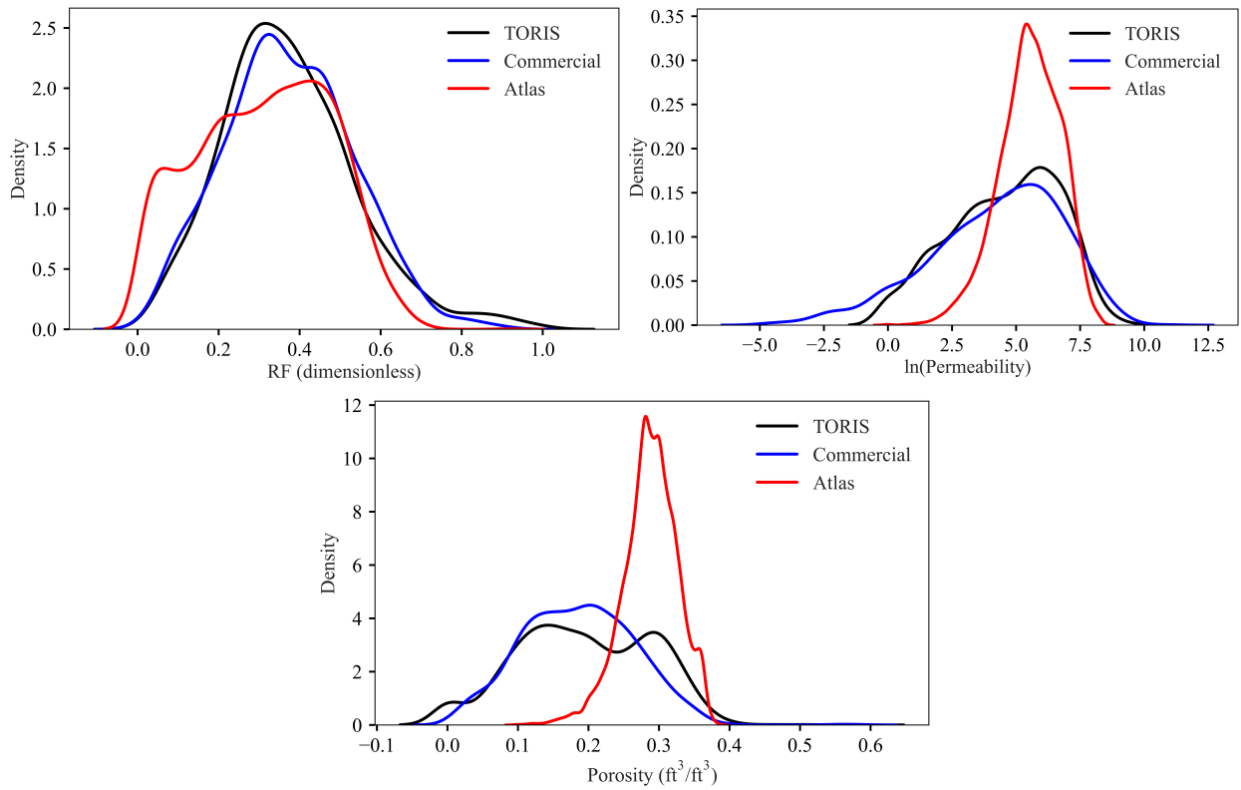

Figure 1. Distributions of the oil RF (dimensionless), porosity (ft$^3$/ft$^3$), and natural logarithm of permeability for databases used in this study.

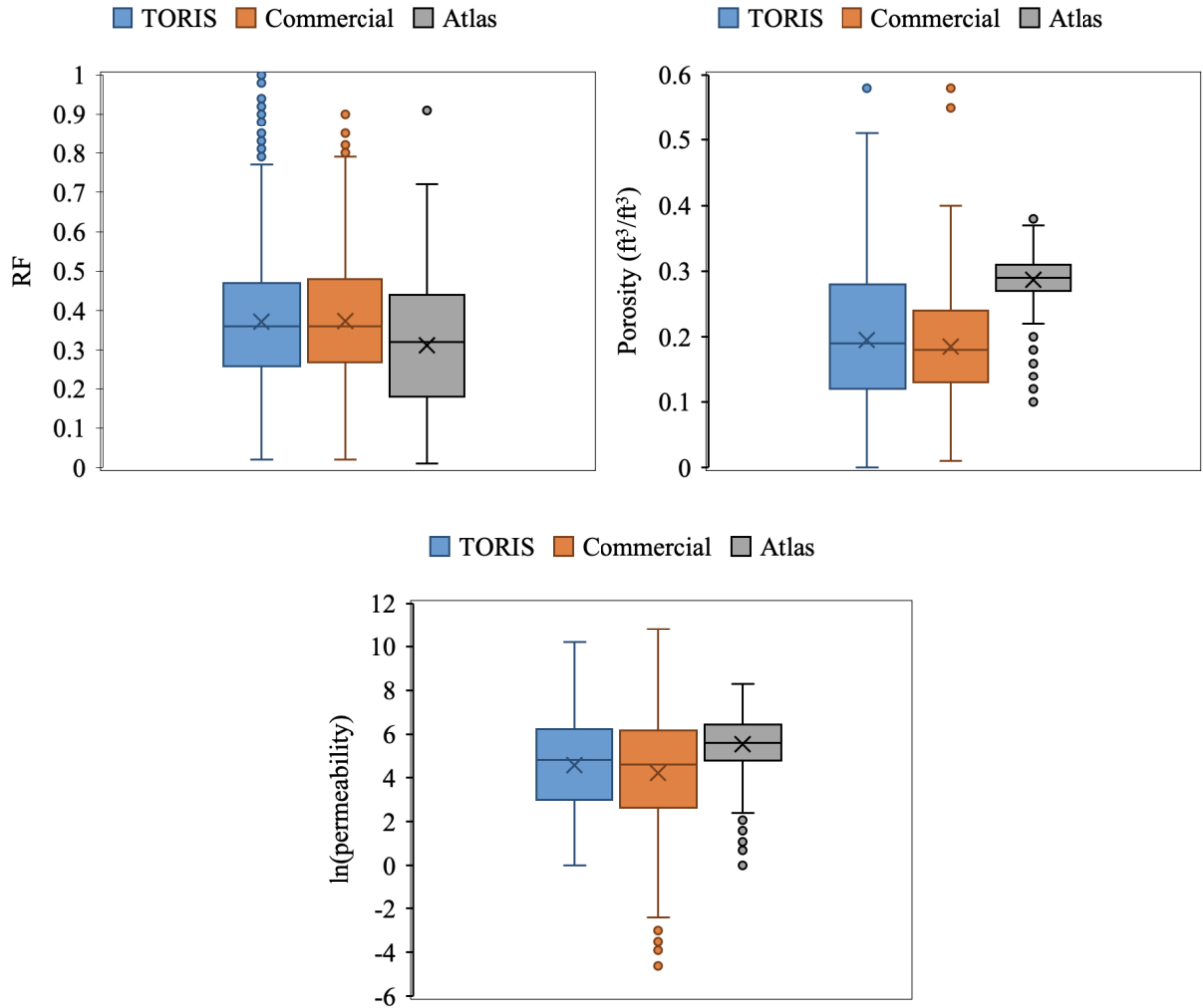

Figure 2. Box and whisker plot of the oil RF, porosity (ft$^3$/ft$^3$), and natural logarithm of permeability for the three databases used in this study. Crosses in each box represents the mean of the distribution. The lines inside the boxes show the median of the distribution. The lower boundaries in each box show the 25$^{th}$ quantile and the upper boundaries of the boxes show the 75$^{th}$ quantile. The horizontal lines at the end of the vertical lines show the smallest and largest data points of the distributions and dots represent the outliers in each distribution.

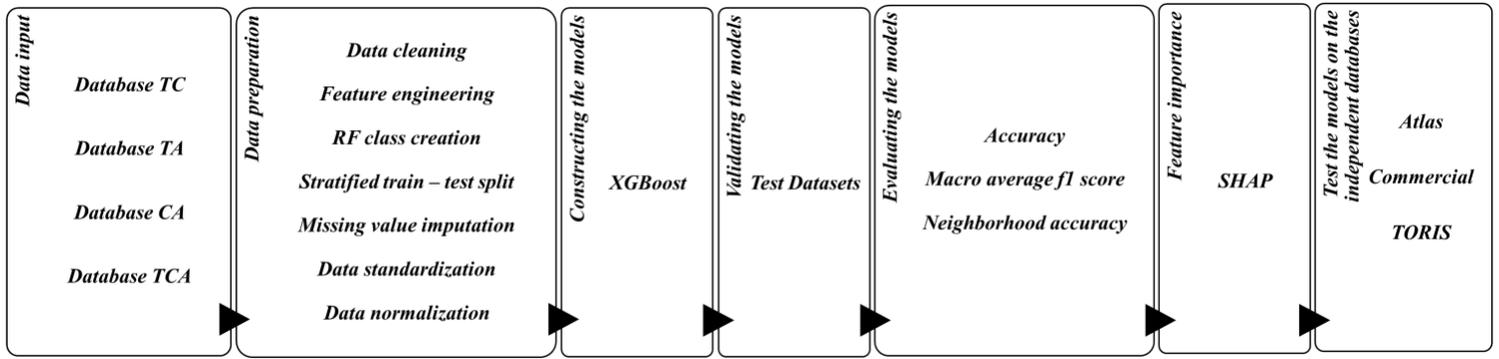

Figure 3. The workflow used to prepare the data, construct the models and evaluate the models on independent databases.

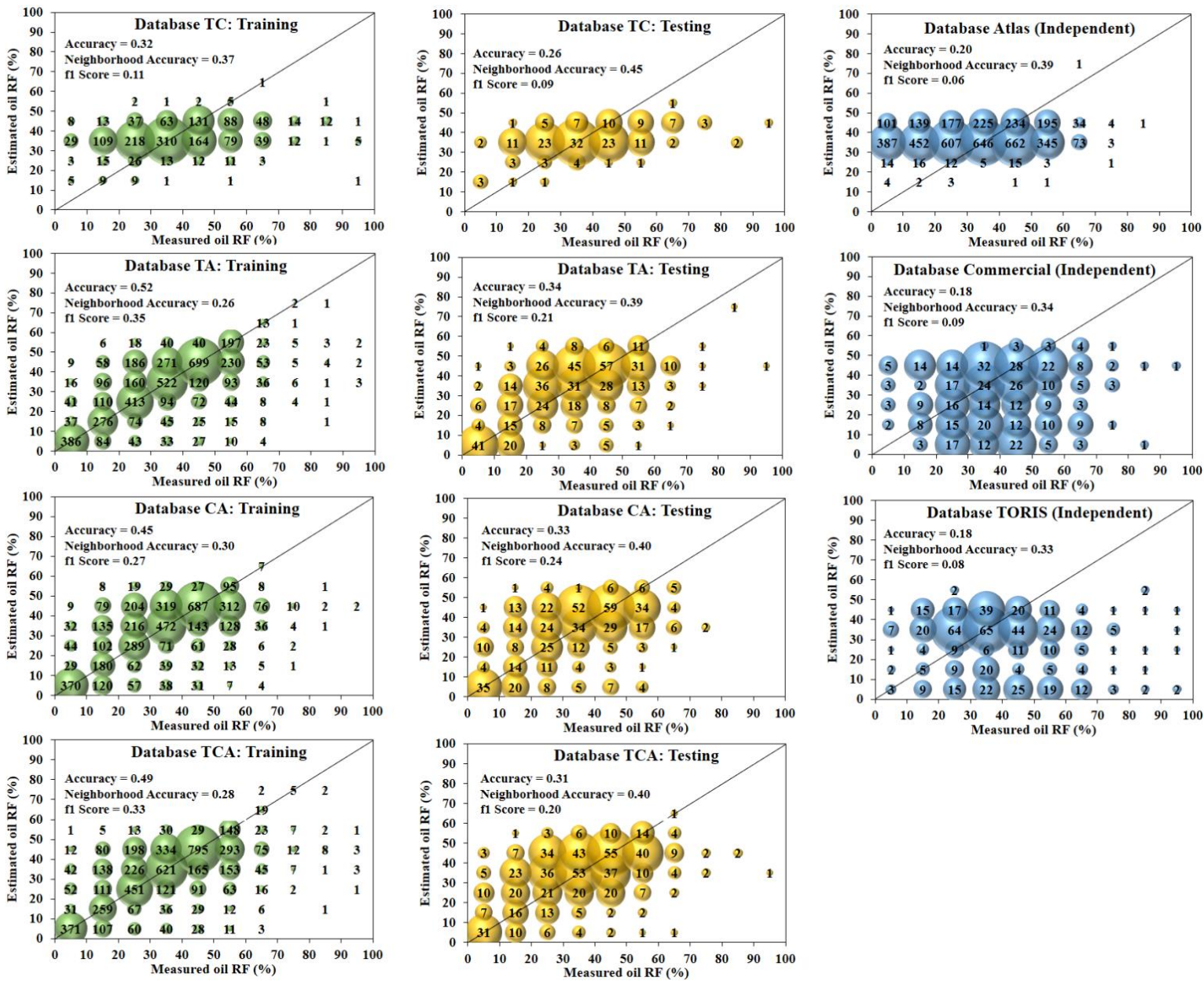

Figure 4. Bubble charts of estimated oil RF classes versus the actual RF classes by the XGBoost classification algorithm for the training and testing datasets in addition to the independent database. The number on each bubble indicates the number of samples in that RF class, and the solid line represents the 1:1 line. TC denotes TORIS merged with Commercial, TA is TORIS merged with Atlas, CA represents Commercial merged with Atlas, and TCA denotes Commercial merged with TORIS and Atlas. The RF values were presented in percentages for better visualization.

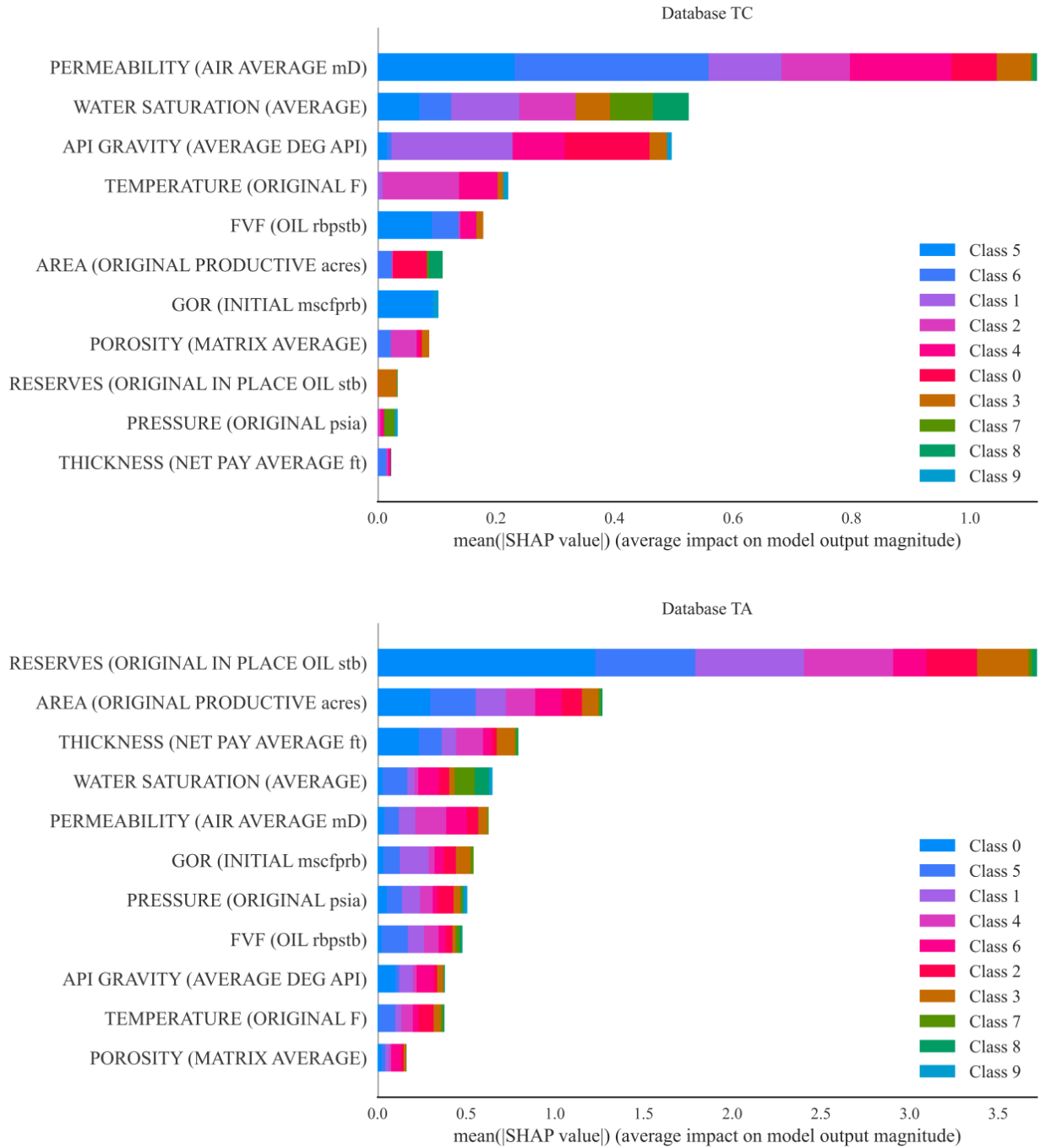

Figure 5. Feature importance and average impact on the model for the XGBoost algorithm in the prediction of RF class for the databases TC (TORIS merged with commercial), TA (TORIS merged with Atlas, CA (commercial merged with Atlas), and TCA (TORIS merged with commercial and Atlas. Each feature's contribution in predicting each class is also shown.

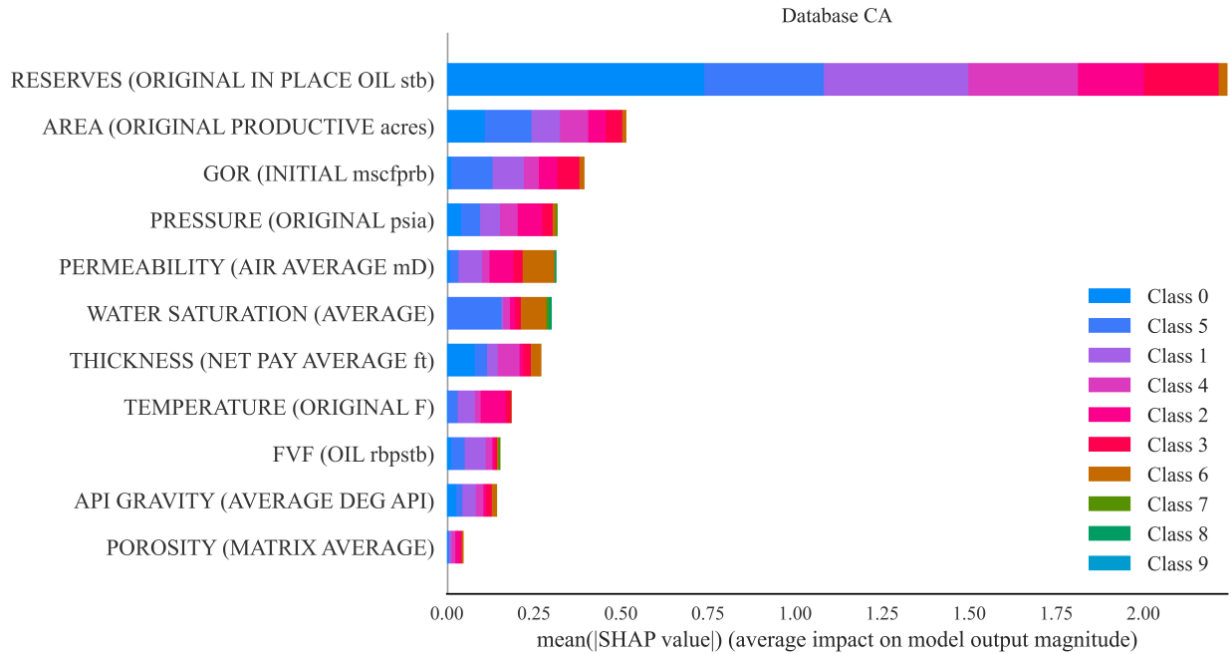
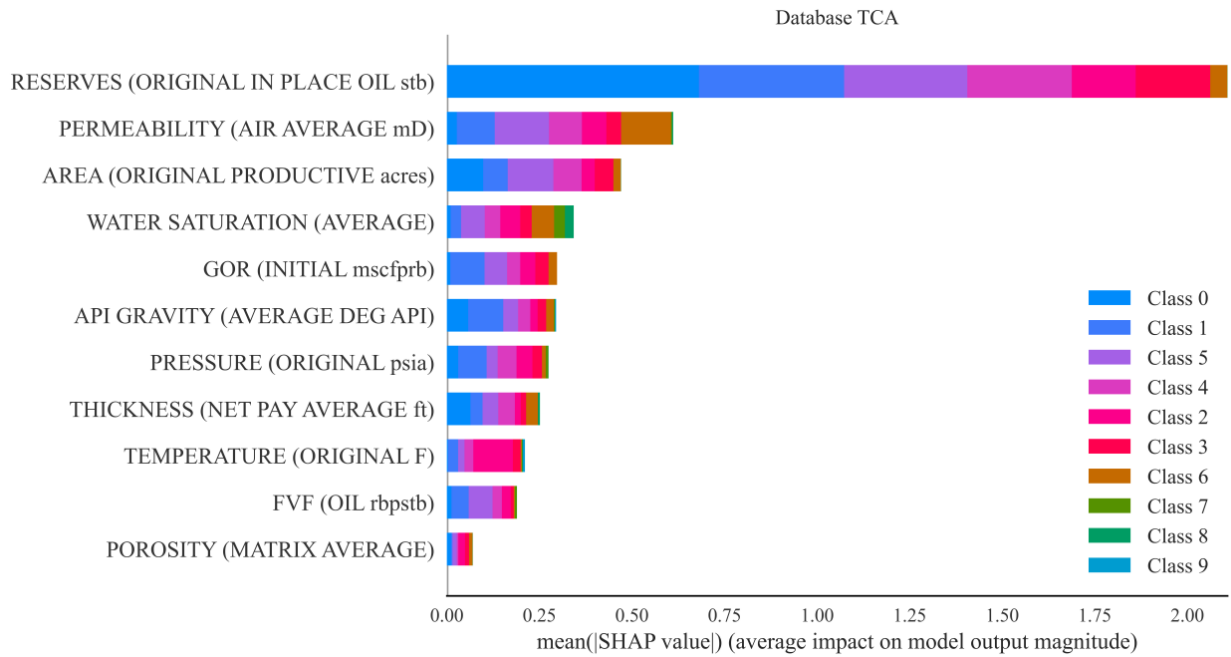

Figure 5. Continued.

Table 1. Final input features and target variable (RF) as well as their ranges within each generated database. Recall that TC is TORIS merged with commercial, TA is TORIS merged with Atlas, CA is Commercial merged with Atlas, and TCA is the combination of TORIS, commercial and Atlas.

| Feature | Database TC | Database TA | Database CA | Database TCA |
|---|---|---|---|---|
| API Gravity | 4-73 | 4-60 | 7-73 | 4-73 |
| $B_o$ (RB/STB) | 0.99-169 | 0.99-3.3 | 1-169 | 0.99-169 |
| GOR* (MSCF/RB) | 0.001-356 | $0.006\text{-}7.73\times10^7$ | $0.001\text{-}7.73\times10^7$ | $0.001\text{-}7.73\times10^7$ |
| Water saturation (-) | 0-0.95 | 0-0.86 | 0.01-0.95 | 0-0.95 |
| Temperature (°F) | 43-390 | 47-433 | 42.8-433 | 42.8-433 |
| Pressure (psi) | 70-16066 | 140-21609 | 70.1-21609 | 70.1-21609 |
| Thickness (ft) | 2-10850 | 1-2300 | 1-10850 | 1-10850 |
| Reserves (STB) | $2.33\times10^6\text{-}9.33\times10^{11}$ | $1\text{-}2.2\times10^{10}$ | $1\text{-}9.33\times10^{11}$ | $1\text{-}9.33\times10^{11}$ |
| Permeability (mD) | 0-5000 | 0-26817 | 0.01-50000 | 0-50000 |
| Porosity ($ft^3/ft^3$) | 0-0.58 | 0-0.58 | 0.01-0.58 | 0-0.58 |
| Area (acre) | $50\text{-}1.73\times10^7$ | 1-218000 | $1\text{-}1.73\times10^7$ | $1\text{-}1.73\times10^7$ |
| Oil RF | 0.02-1.44 | 0.01-2.32 | 0.01-2.32 | 0.01-2.32 |

API: American Petroleum Institute, $B_o$: oil formation volume factor (function of temperature and pressure), GOR: Gas Oil Ratio, MSCF/RB: Thousands Standard Cubic feet per Reservoir Barrel, STB: Stock Tank Barrel, $S_w$: water saturation, $\phi$: porosity, RF: ultimate recovery factor.

Table 2. Hyperparameters and their optimized values for the XGBoost model developed on four combinations of three databases. Recall that TC is TORIS merged with commercial, TA is TORIS merged with Atlas, CA is Commercial merged with Atlas, and TCA is the combination of TORIS, commercial and Atlas. The training datasets in the TC, TA, CA, and TCA databases consisted of 1502, 4779, 4654, and 5461 samples, respectively.

| Parameters | Database TC | Database TA | Database CA | Database TCA |
|---|---|---|---|---|
| Max depth | 2 | 4 | 4 | 5 |
| Minimum child weight | 6 | 3 | 2 | 2 |
| Learning rate | 0.1 | 0.05 | 0.05 | 0.05 |
| Subsample | 0.9 | 0.8 | 0.9 | 0.9 |
| Column sample by tree | 0.9 | 1 | 1 | 1 |
| Objective | Multi:softmax | Multi:softmax | Multi:softmax | Multi:softmax |
| Evaluation metric | mlogloss | mlogloss | mlogloss | mlogloss |
| Alpha | 0.2 | 0.8 | 0.3 | 0.9 |
| Lambda | 0.01 | 0.06 | 0.04 | 0.03 |
| Column sample by level | 0.9 | 1 | 1 | 1 |
| Gamma | 0.01 | 0.01 | 0.01 | 0.01 |
| Max delta step | 0.1 | 0.1 | 0.2 | 0.2 |
| Number of classes | 10 | 10 | 10 | 10 |

Table 3. The values of Accuracy, macro averaged f1 score, and neighborhood accuracy (given in parentheses in the accuracy columns) for the XGBoost classification algorithm employed to predict the oil RF class in the databases used in this study.

| Database | Training | | Testing | | Independent | | |
|---|---|---|---|---|---|---|---|
| | Accuracy | Macro Averaged f1 Score | Accuracy | Macro Averaged f1 Score | Database | Accuracy | Macro Averaged f1 Score |
| TC* | 0.32 (0.37) | 0.11 | 0.26 (0.45) | 0.09 | Atlas | 0.20 (0.39) | 0.06 |
| TA* | 0.52 (0.26) | 0.35 | 0.34 (0.39) | 0.21 | Commercial | 0.18 (0.34) | 0.09 |
| CA* | 0.45 (0.30) | 0.27 | 0.33 (0.40) | 0.24 | TORIS | 0.18 (0.33) | 0.08 |
| TCA* | 0.49 (0.28) | 0.33 | 0.31 (0.40) | 0.20 | - | - | - |

* The training datasets in the TC, TA, CA, and TCA databases consisted of 1502, 4779, 4654, and 5461 samples, respectively.